\documentclass[letterpaper, 10 pt, conference]{ieeeconf}  % Comment this line out if you need a4paper

\IEEEoverridecommandlockouts                              % This command is only needed if 
\usepackage{amsmath,amssymb,amsfonts}
\usepackage{algorithmic}
\usepackage{graphicx}
\usepackage{braket}
\usepackage{textcomp}
\usepackage{xcolor}
\usepackage{subcaption}
\usepackage{float}
\usepackage{multicol}
\usepackage{caption}
\usepackage[colorinlistoftodos]{todonotes}

\usepackage[normalem]{ulem} 

\title{\LARGE \bf
Improving Grasp Stability with Rotation Measurement \\from Tactile Sensing
}

\author{Raj Kolamuri$^{*,1}$, Zilin Si$^{*,2}$, Yufan Zhang$^{1}$, Arpit Agarwal$^{2}$ and Wenzhen Yuan$^{2}$% <-this % stops a space
\thanks{$^{*}$Authors with equal contribution.}%
\thanks{$^{1}$Raj Kolamuri and Yufan Zhang are with the Department of Mechanical Engineering, Carnegie Mellon University, 5000 Forbes Ave, Pittsburgh, PA 15213, USA
        {\tt\small \{rkolamur, yufanzha\}@andrew.cmu.edu}}%
\thanks{$^{2}$Zilin Si, Arpit Agarwal and Wenzhen Yuan are with the Robotics Institute, Carnegie Mellon University, 5000 Forbes Ave, Pittsburgh, PA 15213, USA
        {\tt\small \{zsi, arpita1, wenzheny\}@andrew.cmu.edu}}%
}

\begin{document}

\maketitle
\thispagestyle{empty}
\pagestyle{empty}

%%%%%%%%%%%%%%%%%%%%%%%%%%%%%%%%%%%%%%%%%%%%%%%%%%%%%%%%%%%%%%%%%%%%%%%%%%%%%%%%
\begin{abstract}
% We present a torsion detection algorithm and build a close-loop grasping framework based on tactile sensor, where we detect the failure in the process of grasping with large torsion and use it to guide the re-grasp. 
% A common case of grasping failure is when the object is grasped at the wrong location, causing rotational displacement between the gripper and object. We develop a model-based algorithm to detect the onset of rotation and the degree of rotation from the GelSight sensing, and further use it to guide the regrasp location adjustment until the stable grasp is reached. 
% We conduct offline experiments to test our torsion detection algorithm's accuracy with ground truth and online experiments to proof the efficacy of our system. 

% \arpit{Is this a large number for regrasp tasks? If yes, then the sentence could go like "We validate our grasp-regrasp system on over a 100 grasping attempts.}

Rotational displacement about the grasping point is a common grasp failure when an object is grasped at a location away from its center of gravity. Tactile sensors with soft surfaces, such as GelSight sensors, can detect the rotation patterns on the contacting surfaces when the object rotates. In this work, we propose a model-based algorithm that detects those rotational patterns and measures rotational displacement using the GelSight sensor. We also integrate the rotation detection feedback into a closed-loop regrasping framework, which detects the rotational failure of grasp in an early stage and drives the robot to a stable grasp pose. We validate our proposed rotation detection algorithm and grasp-regrasp system on self-collected dataset and online experiments to show how our approach accurately detects the rotation and increases grasp stability. 
% We perform experiments to show how the proposed algorithm can improve grasp stability by incorporating  

% From more than one hundred times grasping, we showed our system can efficiently finish the grasp-regrasp task.

\end{abstract}

%%%%%%%%%%%%%%%%%%%%%%%%%%%%%%%%%%%%%%%%%%%%%%%%%%%%%%%%%%%%%%%%%%%%%%%%%%%%%%%%
\section{INTRODUCTION}
Robotic grasping is a long-studied problem and the foundation of many manipulation tasks~\cite{grasp-review}. Frequently-asked questions in robotic manipulation include how to grasp an arbitrary object stably, detect grasp failure promptly, and take precautions to avoid such failure. Traditional grasping research focused on detecting grasping locations based on the shape of the objects~\cite{model_grasp_sythesis},~\cite{data-driven_grasp_sythesis} which is typically obtained from vision. However, those methods rarely considered objects' physical properties, such as mass, mass distribution, surface friction, and rigidity. Such properties, which can attribute to significant differences in the physical interaction between robots and the target objects, are essential to decide the optimal grasps. 

\begin{figure}[!htbp]
    \centering
    \includegraphics[width=\linewidth]{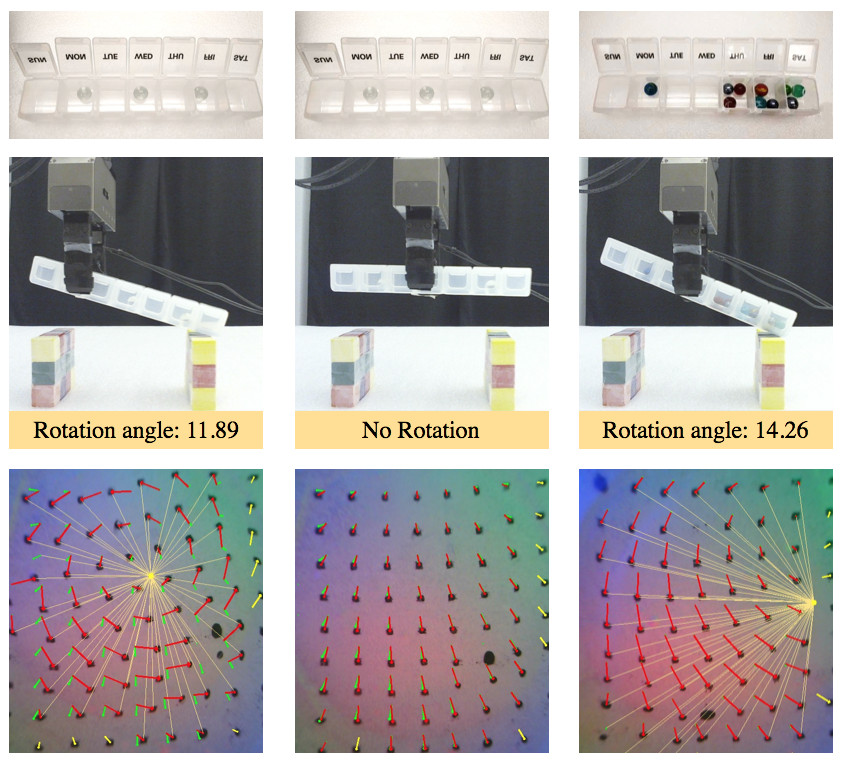}
    \caption{Grasping an object with unknown mass distribution. 
    %Usually, it is hard to tell the mass distribution from vision,
    The center of gravity of an object depends on its mass distribution (Row 1). Grasping away from the center results in contact rotation (Row 2).
    A robot can use tactile images from a GelSight sensor (Row 3) to measure contact rotation caused by torque when lifting an object, and therefore move to a grasp point close to the center of gravity and conduct a stable grasp. 
    %but tactile sensing can provide rich contact information about the grasp. 
    % If a robot grasps an object improperly, significant rotation would happen at the contact surface; with the feedback of tactile sensing, the robot can adjust the grasp location and reach a stable grasp as shown in the middle case.
    }
    \label{fig:overview}
\end{figure}

Tactile sensing provides promising solutions to the challenge. By detecting the contact area and contact force during grasping, tactile sensing gives effective feedback about the grasp outcome, which subsequently can be used to build a closed-loop grasping framework. The study of tactile-sensing-based grasp focuses on detecting and measuring slip ~\cite{robot-human-grasp},~\cite{shear-slip} to choose a proper grasping force to avoid object dropping. However, rotation is another common cause of grasp failure, that has not been well studied. This kind of failure happens when a robot grasps an object at locations far from the object's center of gravity. As a result, the large torque at the contact can make the object rotate or make the grasp vulnerable to external impact. Increasing the gripping force does little to mitigate this type of failure. Instead, this problem can be corrected by choosing other grasp locations closer to the object's center of gravity.  

In this paper, we improve the grasp stability by proposing a model-based method to detect the rotation caused by torque using a tactile sensor. Our method can detect the rotational failure of grasp at an early stage and use tactile information to guide a robot to find a stable grasping location that is close to the object's center of gravity. We apply our method to a vision-based tactile sensor called GelSight~\cite{gelsight-base}, which uses a piece of soft elastomer as the medium of contact and an embedded camera to track the deformation of the elastomer surface. The torque on the contact surface will cause torsion on the elastomer medium, which can be visualized as a rotational pattern of the markers painted on the elastomer surface. We measure the markers' rotation angles and directions to estimate the torque, and show the stability of the grasp against the torsional load.
% In our experiment on XXX grasp samples on XX objects, our method can measure the rotation angle of objects within a standard error of XXX, and predict torsional failure in XX\% of the cases. Using rotation measurement results in a closed-loop re-grasping experiment, the robot could successfully grasp objects in 77 out of 88 cases by moving to a location close to the center of gravity. Our method is simple, yet effective in detecting existing or potential torsional failure and thereby improving grasp stability. 
Because our method is based on the physical feedback during the contact, it applies to a wide variety of arbitrary objects with unknown physical properties, such as mass distribution. With increased grasping stability, our method can help robots operate more robustly and reliably in real-world environments.

\section{RELATED WORK}
\subsection{Grasping and Regrasping}
Computer vision techniques have been employed in robotic manipulation tasks to synthesize grasp poses using visual data of objects and scenes either with model-based methods~\cite{primitive},~\cite{singlepose} or learning-based methods~\cite{predict_grasp},~\cite{novelobjectgrasp}. However, they do not infer the objects' properties, such as mass distribution, friction coefficient, and rigidity. Lack of information of these properties can cause inaccurate or even incorrect grasping operations, which may further result in post-grasp effects like rotation, slippage, and even detachment from the gripper, which may ultimately lead to grasping failure. We use RGB-D data in our work to estimate the geometric center of the object as the initial grasping location and then use a tactile sensor's feedback to adjust the grasp locations.

% We also use 3D point cloud data from RGB-D camera to estimate the geometric center of the object and conduct the first grasp, where we assume in most cases, the geometric center is close to the center of gravity. However, we build upon this using tactile sensor to deal with failure cases.And with the feedback from the tactile sensing, the grasping poses are adjusted with certain policies to search the stable grasping pose.
% \wenzhen{This paragraph can be greatly shortened, since it's not that relevant. You can list a sequence of papers and use one sentence to summarize them.}

Tactile sensors can complement vision data by providing local information at the grasp location, such as measuring slippage~\cite{shear-slip},~\cite{biomim-slip}, force~\cite{vision-touch}, grasp stability~\cite{grasp-stability},~\cite{objPoseSpecificGrasps}, \cite{graspStabilityCNN} and object's Center of Mass (COM)~\cite{weightdist},~\cite{chris-COM}. \cite{predict_grasp} combined tactile sensors with vision sensors to obtain better predictions of grasp outcomes using deep neural networks.
% ~\cite{deformable-obj} used a tactile sensor to estimate slip by tracking change in contact tangential force and counteracting it with increased normal force, thereby preventing grasp failure in deformable objects.
\cite{graspingwithouteye} obtained grasp poses by probing a given workspace and localizing objects from tactile signals without using vision. \cite{analyticGraspSuccess} used tactile feedback to reduce uncertainty in contact position and orientation. However, grasp failure due to incorrect grasp locations is rarely studied using tactile sensors, and we address this problem through our work.

% Center of Mass (COM) is an important physical property of object to help grasping.~\cite{weightdist} used a pressure based tactile sensor and trained an LSTM network to estimate weight distribution of objects.~\cite{chris-COM} used a force/torque sensor and a Newton-Euler formulation of rigid body dynamics to estimate the center of mass location of an object during grasping. Their method required estimating the objects' mass before determining the location of the COM. ~\cite{Center-of-mass-grasp} calculated the COM of an object by defining the weighted distance of the object from a fixed point where weights were proportional to the object's density at the corresponding location from point cloud data. The authors note that this method cannot accurately locate the COM for objects of non-uniform mass distribution.

Regrasping has been used to improve grasp success rate by correcting grasp characteristics, such as pose and force, using feedback from additional sensors. 
% ~\cite{In-hand_proximity} learned a grasp stability predictor and then trained a regrasping algorithm to improve grasp success rate using reinforcement learning techniques.
In~\cite{tactile_regrasp}, the author proposed a method to simulate tactile images at different locations of the object from an initial grasp point, predict grasp success for each of these images, and choose the corresponding location with the highest predicted score for regrasping.~\cite{Center-of-Mass} used a low-resolution pressure tactile sensor to detect grasp failure due to slip, and implemented an LSTM model trained on both tactile and force/torque data to sample a regrasp location achieving stable grasp.~\cite{Center-of-mass-grasp} solved a similar problem like ours to detect wrist moments when large objects are grasped at locations away from their COM. The aforementioned work used both force and torque information from a force/torque sensor to detect grasp instability and guide the robot towards the object's COM. In our work, instead of using force/torque information, we detect similar instabilities in smaller household objects using only images from a tactile sensor. 
% ~\cite{Center-of-mass-grasp} combined 3D range and force/torque sensing to find the center of mass and reached the efficient handle grasp. We also form a closed-loop regrasp framework to improve the grasp stability.

\subsection{Slip Detection based on Tactile Sensing}
% \wenzhen{This paper is not about 'slip detection'. Then why do you review those many works on slip detection? You need to explain}
One of the primary motivations of our work is the capability of tactile sensors to detect slippage. 
% Slippage while manipulation leads to a classical grasp failure. 
% Detecting slip before or during grasp and counteracting 
% Tactile sensors have been explored to detect, classify and measure slip. 
Several model-based~\cite{review},~\cite{piezoelectret},~\cite{BioTac} and learning-based~\cite{calib_slip},~\cite{learn_slip_detect} methods have used tactile sensors to detect, classify and measure slip. 
% Slip can be classified into two categories: translational slip and rotational slip. 
Rotational slip can also result from an incorrect grasping location.  Some early works such as~\cite{3-axis_rotation_slip} and~\cite{tactile_force} used force/torque tactile sensors to classify translational and rotational slip by analyzing contact area.~\cite{Biomimetic} and~\cite {Tactile_CNN_rotation} trained neural networks to classify linear and rotational slip from force/pressure sensor data and piezo-resistive sensor data. 
\cite{ContactCentroid} discussed the concept of a contact centroid and proposed a method to obtain rotational spin about this point using remote force/torque sensor readings.

The GelSight sensor is a vision-based tactile sensor that provides high-resolution information of the contact region. 
% It consists of a soft elastomer gel illuminated with red, green, and blue lights from different directions. The gel is painted with black markers in a rectangular grid. A camera is planted in the gel that captures the gel surface as a 2D RGB image. 
 When an object comes in contact with the gel, the gel surface deforms, and the markers on the gel undergo radial motion. By tracking these marker motions, users can study local phenomena like slippage and rotation in great detail. 
There have been some related studies~\cite{shear-slip},~\cite{Improved_Gelsight} to detect the occurrence of slip using Gelsight tactile sensors, but they are restricted to numerically identifying slip. 
% Moreover, ~\cite{tactile_vision} presented a slip detection method by fusing vision and tactile sensing. The authors trained a DNN to classify a grasp is stable or not.
% ~\cite{in-hand} obtained in-hand pose estimate of objects using Deep Gated Multi-modal Learning with inputs from the GelSight sensor. 
We use the GelSight sensor in our analysis to track and analyze the rotational patterns of objects at the contact location. To the best of our knowledge, this is the first attempt at detecting rotation of objects and measuring rotation angle and its orientation using only a vision-based tactile sensor.

\section{METHOD}

When grasping an object at a location away from the object's center of gravity, the object undergoes rotation about the gripping point due to the moment applied by gravitational force. We show that it is possible to detect this rotation and measure it in real-time using images from the GelSight sensor. The GelSight sensor outputs tactile images with markers labeled on them. These markers trace the object's movement when the sensor is in quasi-static contact with the object, hence providing high-resolution localized information in the contact region. When rotation occurs, it gives rotational patterns of markers on the GelSight surface. We analyze these patterns to determine the Center of Rotation (COR) and eventually calculate the rotation angle. We then show how this method can help a robot to reach a stable grasp pose.

\subsection{Contact Rotation Measurement Overview}
% \begin{figure}[!htbp]
%     \centering
% `    \includegraphics[width=\linewidth]{imgs/rotation_pipline.png}
%     \caption{rotation detection flow chart. After the tactile sensor gets in touch with the object, our algorithm detects the stable contact, whether the rotation happens and quantifies the rotation.}
%     \label{fig:rotation_pip}
% \end{figure}

% We propose a model based algorithm to detect rotation and measure both its magnitude and orientation by processing the time sequence of the GelSight images. When rotation occurs due to wrong grasp conditions, the markers on the GelSight image exhibit rotational patterns as they trace the motion of the object’s part that is in contact with the GelSight sensor. The algorithm analyzes these rotational patterns to quantify the rotation of markers and also give its orientation(CW/CCW). 
Our algorithm proceeds through three steps: 1) Object Contact Detection: Initially, the system keeps looping to check the occurrence of contact between the sensor surface and the object. 2) Detection of Rotation Onset: It then tracks the markers on the GelSight image and keeps checking if rotational patterns occur on the surface. 3) Rotation Angle Measurement: If rotation starts, it then analyzes the marker motions and calculates the Center of Rotation(COR). The algorithm then gives a measure of the angle of rotation and its orientation about the calculated COR. The following subsections detail each step of the algorithm. 

\subsection{Object Contact Detection}
Precise contact detection is an essential pre-processing step for rotation detection. When a grasp occurs, the markers that contact the object can directly trace the objects' motion. In the subsequent frames, these markers, designated as contact markers, can indicate whether the object undergoes rotation or remains stable. The remaining markers are designated as non-contact markers.

The grasping process takes time to complete and the contact area on the gel is dynamic. We first identify a stable contact state of the grasp and then the contact region in the image corresponding to it. To accommodate different situations, we determine a stable grasp state in two ways: soft stable contact and hard stable contact. After 10 frames from the start, if the change in the frame-frame markers' motion magnitude is less than a set threshold, we say the contact is a soft stable contact. However, in some cases, there is hardly a stable contact stage and we designate the frame after 30 frames($\sim$1sec) as the hard stable contact.
% The contact area is dynamic during the grasp; therefore, we wait until the contact becomes stable to detect the contact region. We set two timestamps, one as soft stable contact and the other as hard stable contact, to accommodate different situations. After 10 frames from the start of the contact, if the change in markers' motion magnitude from the last frame is less than a small threshold, we say the contact is stable, namely, a soft stable contact. However, in some cases, there is hardly a stable contact stage in which we set 30 ($\sim$1sec) frames, a hard time stamp, as stable contact detection after the start of contact. 

\begin{figure}[t]
    \centering
    \includegraphics[width=\linewidth]{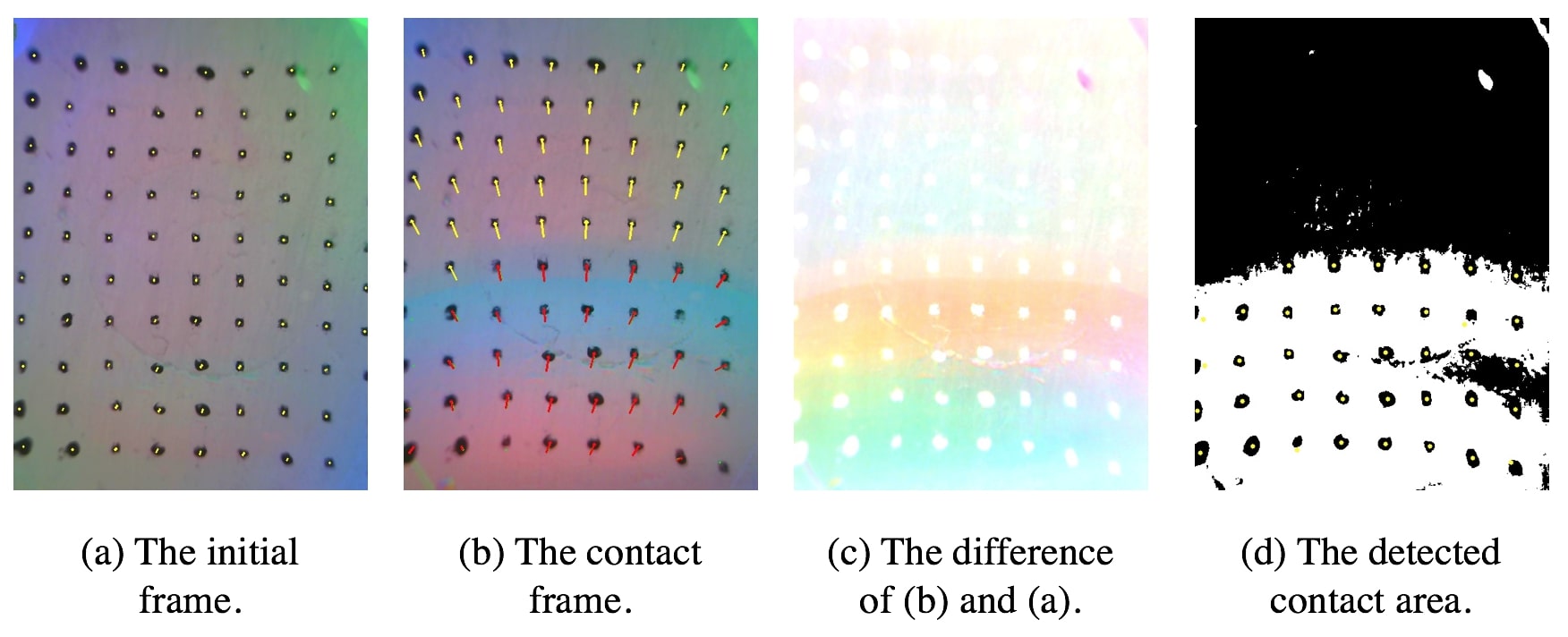}
    \caption{Contact area detection.
    Comparing (b) with (a), contact happens mostly in the bottom part of the image as the difference shown in (c), and (d) is our detected contact area.
    }
    \label{fig:contact}
    \vspace{-5mm}
\end{figure}

By the time the contact becomes stable, the illumination changes across the three color channels as seen in Fig.~\ref{fig:contact}(a) and Fig.~\ref{fig:contact}(b). We take the maximum of illumination change across R, G, and B channels at all pixel locations, as in Fig.~\ref{fig:contact}(c), and then filter the resulting image to get the contact region as in Fig.~\ref{fig:contact}(d).
% Meanwhile, we obtain the contact markers with thresholded radial motion in the contact region.

\subsection{Detection of Rotation Onset}
After detecting contact, the algorithm tries to identify the markers' rotational patterns in the subsequent frames. If no such patterns are detected, the grasp is considered stable. Otherwise, we process the patterns to measure rotational displacement.

\begin{figure}[t]
    \begin{subfigure}{0.45\linewidth}
        \includegraphics[width=\linewidth]{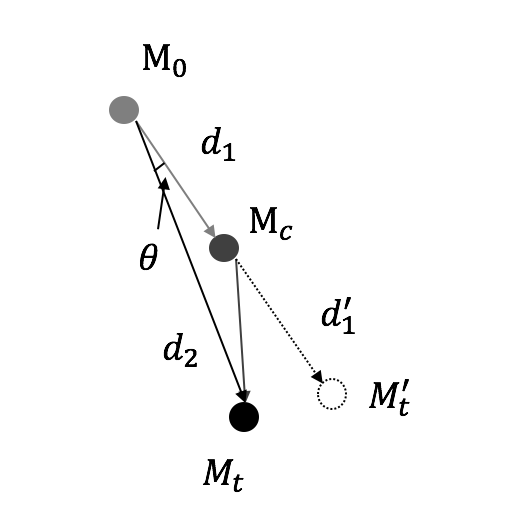}
        \caption{Rotation onset detection}
    \end{subfigure}
    \hfill
    \begin{subfigure}{0.45\linewidth}
        \includegraphics[width=\linewidth]{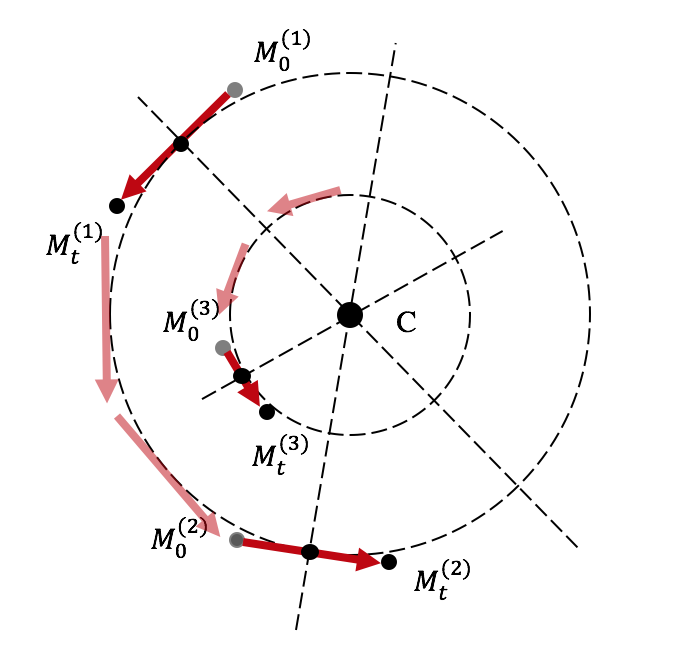}
        \caption{Rotational pattern of marker motions with their COR}
    \end{subfigure}
    \caption{(a) Detection of rotation onset: the angle difference $\theta$ between the motion vector $d_1$ (marker motion from initial frame $M_0$ to the contact frame $M_c$), and the motion vector $d_2$(marker motion from initial frame $M_0$ to the current frame $M_t$) indicates the rotation onset. (b) Detection of center of rotation: the normals of rotating motion vectors intersect at the center of rotation.}
    \label{fig:rotation_onset}
    \vspace{-2mm}
\end{figure}

We calculate the relative motion change and angular change of the marker vectors w.r.t contact markers in any given frame after contact. If either of these changes crosses a fixed threshold, we say that the object is in rotation. The object's translational motion will be considered separately in the following sections.
For instance, in Fig.~\ref{fig:rotation_onset}(a), for a certain marker, we check the value of angle $\theta$ between the motion from the initial frame $M_0$ to the stable contacting frame $M_t$ as $d_1$ and the motion from the initial frame $M_0$ to the current frame $M_c$ as $d_2$. We check the magnitude of the motion from the stable contact frame to the current frame as well. If one of them goes beyond a threshold, we consider that as the onset of rotation, and start to measure the rotation angle.
% \vspace{-5mm}

\subsection{Rotation Angle Measurement}
In this step, we detect the location of the center of rotation (COR) and measure rotational displacement about this center. We take the motion of markers from the initial frame and draw normals to each of them. The intersection point of these normal vectors would ideally be the COR. However, due to noise, they might not intersect at a single location. Hence, we formulate a least-squares solution to obtain the best estimation of COR. 

Illustrated in Fig.~\ref{fig:rotation_onset}(b), assuming the COR is $C = (x_c,y_c)$, and one motion vector is from $M_0 = (x_0,y_0)$ to $M_t = (x_t,y_t)$, then the motion vector is perpendicular to the direction joining the COR and the mid point of the motion vector $X_m = (x_m,y_m) = ((x_0+x_t)/2,(y_0+y_t)/2)$. Under the ideal situation with no sensor noise, the perpendicularity can be satisfied as

\begin{equation}
\begin{split}
    & (x_c - x_m, y_c - y_m) \perp (x_0 - x_t, y_0 - y_t) \\
    \Rightarrow & (x_c - x_m)(x_0 - x_t) + (y_c - y_m)(y_0 - y_t) = 0 \\
    \Rightarrow & x_c + \frac{y_0 - y_t}{x_0 - x_t} y_c = x_m + \frac{y_0 - y_t}{x_0 - x_t} y_m
\end{split}
\end{equation}
Therefore, the best estimation of COR $\hat{C} = (\hat{x_c}, \hat{y_c})$ can be solved with a set of motion vectors by forming in normal equation $A x = b$ as
\begin{equation}
\begin{split}
   & \left[\begin{matrix}
   1 & \frac{y^{(1)}_0 - y^{(1)}_t}{x^{(1)}_0 - x^{(1)}_t} \\
   1 & \frac{y^{(2)}_0 - y^{(2)}_t}{x^{(2)}_0 - x^{(2)}_t} \\
   ... \\
   1 & \frac{y^{(n)}_0 - y^{(n)}_t}{x^{(n)}_0 - x^{(n)}_t}
  \end{matrix} \right] 
  \left[\begin{matrix}
  x_c \\
  y_c
  \end{matrix} \right] 
  = \left[\begin{matrix}
   x^{(1)}_m + \frac{y^{(1)}_0 - y^{(1)}_t}{x^{(1)}_0 - x^{(1)}_t} y^{(1)}_m \\
   x^{(2)}_m + \frac{y^{(2)}_0 - y^{(2)}_t}{x^{(2)}_0 - x^{(2)}_t} y^{(2)}_m \\
   ... \\
   x^{(n)}_m + \frac{y^{(n)}_0 - y^{(n)}_t}{x^{(n)}_0 - x^{(n)}_t} y^{(n)}_m
  \end{matrix} \right] \\
\end{split}
\end{equation}
And the rotation center C is estimated by 
\begin{equation}
    x = (A^{T}A)^{-1}A^{T}b
\end{equation}

The angle of rotation is calculated as the angle $\theta$ of\\
 $<\overrightarrow{X_c M_{0}},\overrightarrow{X_c M_t}>$:
\begin{equation}
    \theta = arccos(\frac{<\overrightarrow{X_c M_{0}},\overrightarrow{X_c M_t}>}{|\overrightarrow{X_c M_{0}}||\overrightarrow{X_c M_t}|})
\end{equation}

However, the angle calculated for each maker is an absolute rotation angle value, which does not have direction. We calculate the orientation of rotation by taking the moment of the motion w.r.t the COR. If the sign of the moment is positive, it is counter-clockwise and clockwise if negative. 

The nature of the gel and its interaction with the objects can result in noise with some markers exhibiting orientation in the opposite direction to the object's rotation direction. Therefore, we calculate the moments about COR for all the markers and do a majority vote to decide the final orientation of rotation. If a rotation occurs, one orientation group should dominate the others. However, if an object is grasped and remains static, it could be recognized as a rotation case due to the markers' radial motion. In such a scenario, the number of markers in both groups are similar, which can distinguish false-positive cases from the real rotation cases with dominant numbers as described above.

\begin{figure}[t]
    \centering
    \includegraphics[width=\linewidth]{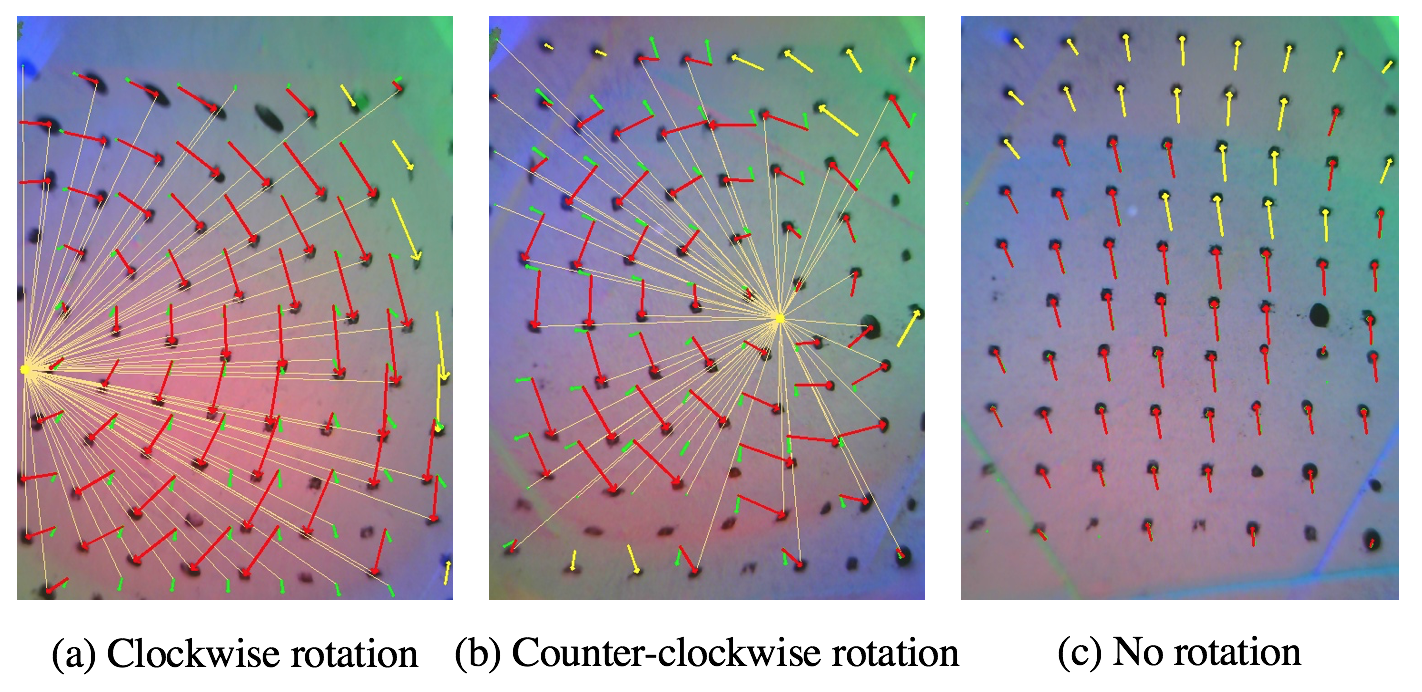}
    \caption{Three types of contact rotation cases which can be measured from the rotational patterns of the black markers in the GelSight images.}
    \label{fig:COR}
    \vspace{-2mm}
\end{figure}

Clockwise rotation and counter-clockwise rotation cases can be seen in Fig.~\ref{fig:COR}(a) and (b), where we find the CORs for contact markers. A grasp can be considered stable in two cases: 1) no rotation of the object occurs, 2) low rotation with small rotation angles, set as 5$^{\circ}$ in our experiments.

\subsection{Special Cases}
\subsubsection{Translational Displacement}
One special case is the translational displacement during grasping. When the gripper lifts the object, the object undergoes translational movement due to the imbalance of friction and gravity if there is insufficient normal contact force. It is natural to increase the normal force~\cite{shear-slip} to avoid this displacement. Markers' motion are mostly similar in direction for translation and dissimilar for rotation. We utilize this fact to distinguish between translational and rotational displacement in our analysis. We apply SVD decomposition on the motion of markers. If there is only one dominating direction, which reflects in only one large singular value, we say the motion is translation.

\subsubsection{Small Contact Area}
\begin{figure}[t]
    \centering
    \includegraphics[width=\linewidth]{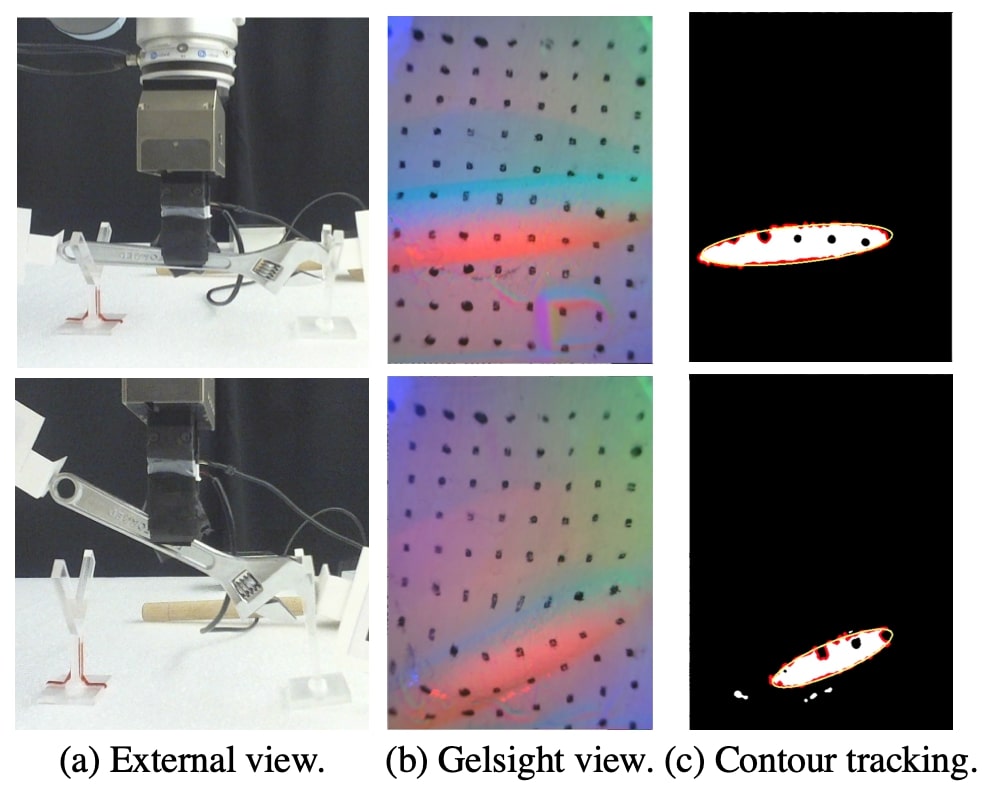}
    \caption{Rotation detection on contacting with irregular area. (a) The case of grasping a wrench. (b) Gelsight image for grasping. (c) We track the the contact area's long axis of the contour.}
    \label{fig:contour}
    \vspace{-2mm}
\end{figure}

Another special case is when the contact happens only in a small area with objects that have irregular shapes. As seen in Fig.~\ref{fig:contour}(b), the contact area is not flat but small with rich geometry, so it is hard to track markers inside the contact area. Alternatively, we measure the rotation by locating and tracking the contact area's contour. Specifically, we track the relative angular change of the contour's principal axis because the rotation of the area can be approximated by the rotation of the principal axis. Based on the construction of the tactile sensor, the lights come from the surrounding of the gel pad. So the contact area with huge gradients' color value is more intense. We convert the RGB format into HSV format, filter and smooth the Value channel to extract the main contact area as shown in Fig.~\ref{fig:contour}(c).

% Another special case is the contact happens only in small area when the objects have irregular shapes. As seen in Fig.~\ref{fig:contour}(b), the contacted area is not flatten but small and with rich geometry, so that it is hard to track the markers inside the contact area. Alternatively we track the rotation by locating and tracking the contact area. Here we extract the contour of the main contact area, and track the motion of the contour to estimate the rotation. Specifically, the angle changing of the direction of the contour's principal axis is the rotational displacement because the rotation of the area can be simplified as the rotation of the principal axis. Based on the construction of tactile sensor, the lights come from the surrounding of the gel pad. So the contact area with huge gradients' color value is more intense. We transfer the RGB color channel to HSV color channel, and threshold and smooth on the Value channel to extract the main contact area as shown in Fig.~\ref{fig:contour}(c).

\subsection{Closed-loop control based on rotation measurement}

We build a regrasp control framework to utilze the GelSight feedback to find a stable grasp position of an object. We setup an RGB-D camera to estimate the object's size as a reference for regrasping adjustment step size.
%which is used as a heuristic parameter to adjust the step size for regrasping. And it is incorporated into a closed-loop control algorithm to search for the stable grasp pose on the object based on the rotation measurement feedback. 
%The following two sections talk about how to estimate the object's size as well as how to control grasp-regrasp to reach a stable grasp at the object's center of gravity.

\subsubsection{Object Size Estimation}

\begin{figure}[t]
    \centering
    \includegraphics[width=\linewidth]{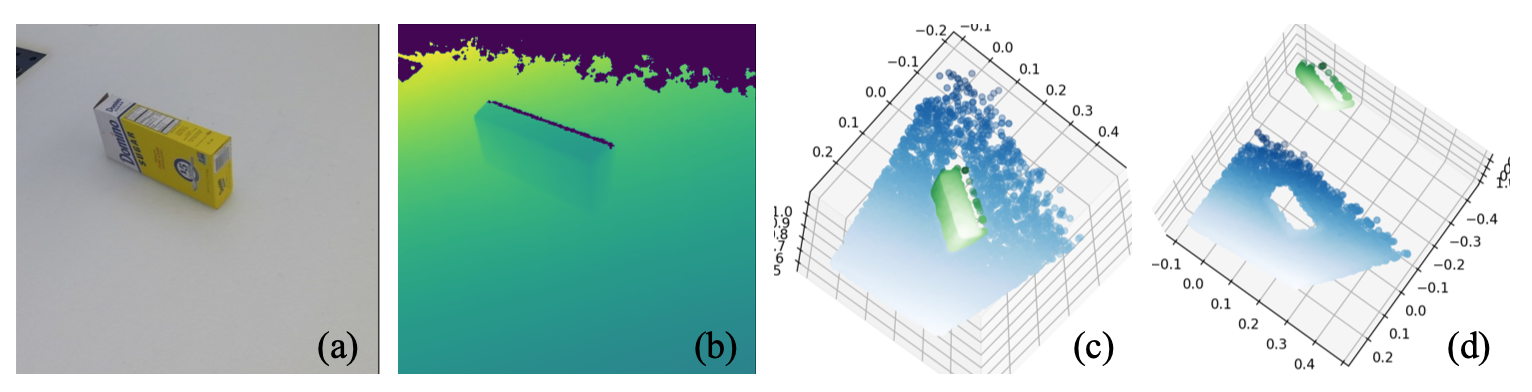}
    \caption{Object length measurement from RGB-D data. The 3D point cloud can be generated from a pair of RGB-D images such as (a) and (b). (c) shows the segmentation of table plane (in blue) and object (in green). We then estimate the object's length from the length of the principal axis of 2D projected point cloud of object on the table plane}
    \label{fig:obj}
    \vspace{-2mm}
\end{figure}

We use the real-scale point cloud generated from the RGB-D data to estimate the object's size, as shown in Fig.~\ref{fig:obj}. Given our experimental setting, where the background is a clean and flat table plane, the plane and the object can be segmented out with RANSAC simultaneously. We only estimate the length of the principal axis along the object because the gripper moves along that axis. To find the principal axis, we project the 3D point cloud of the object to the 2D table plane found from the RANSAC, and apply SVD decomposition to find the principal axis with the largest singular value which is the long axis. We take the 95\% longest distance to the center point as the half of length to tolerate the potential outliers in the point cloud.
\subsubsection{Closed-loop Control} \label{closedloopcontrol}

% With the knowledge of the length of the object, we are able to find the geometry center of the object, and apply the first grasp, since for even distributed-mass object, the geometry center is usually the center of gravity. However, it is unable to predict the mass distribution purely from the vision. If the first grasp is not on the center of gravity, the rotational displacement will happen and we have to regrasp the object based on the tactile rotation measurement feedback. We fix the movement of the gripper along the long axis of the object. If the grasp is detected as unstable, the gripper goes forward and regrasps the object with clockwise rotation detection, otherwise backward. The step size decreases with a certain factor $\alpha$ with more grasping applied where $\alpha < 1$. We keep regrasping the object with the adjusted step size till it reaches a stable grasp. We also try different initial positions away from the midpoint and test the regrasping framework.

With the object's length $L$ measured from vision, we design the regrasp algorithm as following. First, the robot starts with an initial grasp near the geometric center of object. If that grasp is ascertained as rotation, the robot will release the object and move along the object's principal axis to the updated grasp location. The initial regrasp step size is $0.4L$, and then reduced to $\frac{1}{6}L$ for the later regrasps. The initial and adjusted step sizes are heuristic but this configuration is 
verified to be effective in the experiments where the robot could reach a stable grasp pose in a few steps. If the rotation orientation changes during two regrasps, it means that the robot has passed the Center of Gravity. The robot then takes the updated step in the backward direction, and we know that the Center of Gravity is between the two grasp locations. We will then further reduce the step size by a factor of 0.5 for a finer adjustment. 
With this coarse-to-fine grasp adjustment design, our algorithm can quickly find the Center of Gravity within a few grasp trials.
% We build a regrasping framework to utilize GelSight sensor’s feedback to find a stable grasp pose of an object. For each object we start at one edge of the centroidal axis and grasp the object with incremental steps along the axis. The robot grasps the object at each grasp location, and in parallel, the GelSight sensor analyzes each frame according to the above proposed rotation detection algorithm. If the rotation is detected, the robot will place the object back, go to next grasp location and regrasp the object. If no rotation detected, we consider it as a stable grasp and terminate the grasping.

% The robot grasps the object at each grasp location and attempts to lift the object to a maximum height of 5cm. In parallel, the GelSight sensor analyzes each frame according to the above proposed rotation detection algorithm. We designate a grasp to be stable if it doesn't rotate or if the angle of rotation is within 5 degrees. 
% If the algorithm detects the grasp to be unstable during the lifting phase, it raises a preempt signal to the robot through ROS.
% The robot then places the object back. If the GelSight sensor doesn’t detect rotation until 5 cm, the robot waits for 3s to check for rotation detection. If rotation is not detected at this stage, the robot further lifts the object by 3cm and holds it for 5s to indicate that stable grasp pose has been achieved. The grasping terminates once the stable grasp pose is achieved. 

\section{EXPERIMENTS}
\begin{figure*}[t]
\includegraphics[scale=.52]{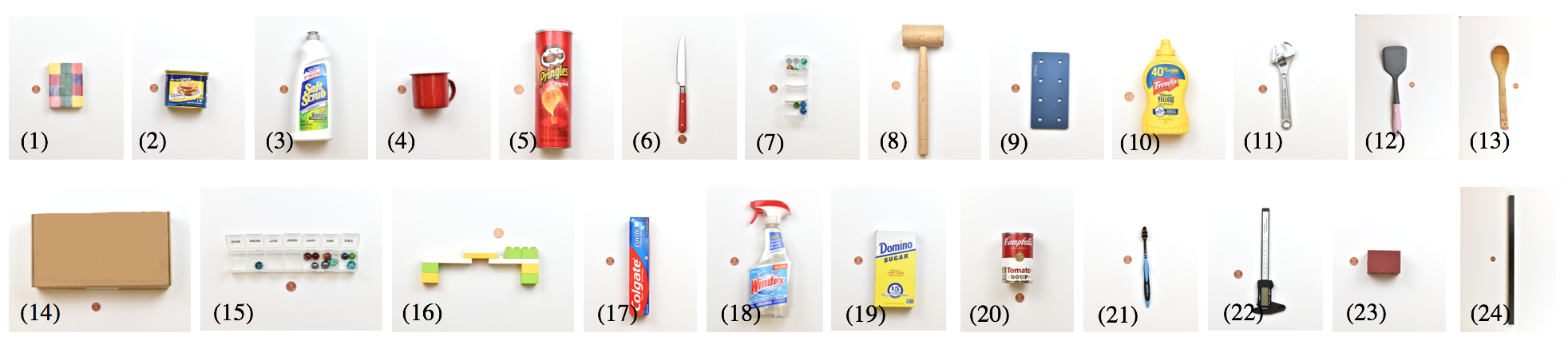}
\caption{Experimental objects with different shapes, mass distributions and materials. For effective evaluation, we include objects of irregular shapes like spatula, wrench, etc. as well as soft objects like sponge bar, tooth paste, etc.} 
\label{fig:objects}
\end{figure*}

We conduct two sets of experiments: offline grasp experiments on contact rotation detection, and closed-loop regrasp experiments based on the rotation measurement results. In the offline experiments, we grasp and lift different objects at different points in a  dataset and then evaluate our rotation measurement algorithm by comparing against the ground truth. In the closed-loop regrasp experiments, we integrate the feedback of the GelSight sensor to update grasp locations until a stable grasp is attained. 

\begin{figure}[!htbp]
    \centering
    \includegraphics[width=\linewidth]{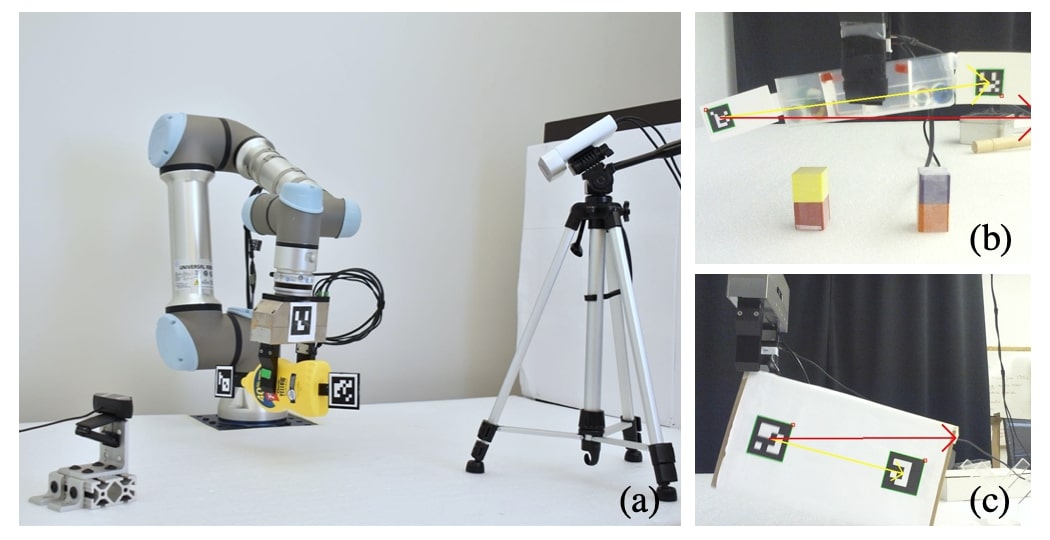}
    \caption{(a): Experimental Setup. A Gelsight tactile sensor is mounted on the gripper, an RGB-D camera is used to estimate the size of object and an external camera on the side measures the ground truth of object rotation from the Aruco Markers attached to the objects. (b) and (c): Ground truth rotation measurement from the external camera. The orientation of Aruco markers in the initial frame and current frame are marked as red and yellow arrows respectively.}
    \label{fig:robot}
     \vspace{-2mm}
\end{figure}

% \subsection{Experiment Setting}
We set up a 6 DOF robot arm (UR5e by Universal Robots) and a parallel jaw gripper (WSG50 by Weiss) to perform grasps. We mount a GelSight sensor on one side and a 3D printed finger(PLA) on the other side. An Azure Kinect camera faces the object from a top view. We then place an external USB camera on the object's side view to obtain ground truth of objects' rotation angles and orientation. The setup is shown in Fig.~\ref{fig:robot} (a).

\subsection{Contact Rotation Measurement on Offline Data}
\textbf{Dataset Collection: } 
We collect data about the rotation of objects to test the rotation detection algorithm's performance. Data includes time-synchronized image sequences from both the GelSight sensor and an external USB camera. The USB camera is placed on the object's side view to capture the Aruco markers and further measure the rotation of the objects as the ground truth. As illustrated in the Fig.~\ref{fig:robot} (b) and (c), two Aruco markers are attached parallel to the object's long axis. Under this setting, the rotation is 2D on the plane of the Aruco markers. This is consistent with the rotational displacement detected from the Gelsight, which is attached on a parallel gripper where only 2D rotational displacements take place in between. Given the real size of the Aruco markers and intrinsic parameters of the camera, the pose of Aruco markers can be estimated using the Perspective-n-Point (PnP) algorithm. We set the initial direction from one marker to another in the first frame, and then calculate the direction change on the Aruco markers' plane as the rotation ground truth for the following frames.
If the algorithm detects rotation and the angle measurement is above 5$^{\circ}$, we consider it an unstable grasp.

We perform 142 grasps on the first 14 objects, as seen in Fig.~\ref{fig:objects}, at multiple locations with gripping speeds varying from 30-50 mm/s and a gripping force of $\sim$30 N. The robot goes uniformly over different locations on the objects along their centroidal axis and grasps at each point. At each grasp, the robot lifts the object upwards to a height of 5 cm and places it back. While the robot lifts the objects, the GelSight sensor and the external USB camera record images at 30 fps. 
% The USB camera is placed parallel to the plane of the GelSight sensor's axis to capture planar rotations of objects. Each object is attached with Aruco Markers, generally at two different ends. When an object rotates, these two markers trace the object's motion. We then measure the rotation of the objects in the world frame by tracking the two markers' orientation change. 
In total, we collected 142 grasp data points with 98 of them labeled as rotational cases and 44 labelled as stable grasps, including translational slip cases using Aruco markers' relative positions. 
% Among 140 grasp data points, rotation was detected \wenzhen{`detected' is weird. It sounds like this is the detection results from your algorithm. If you meant the groundtruth label, you should make it clear these are rotation cases}in 99 cases, and in the rest of 41 cases, the objects were either grasped successfully or with translational slip, which we do not consider as rotational failure.    
% \caption{Rotation onset time delay in  offline experiments. The absolute time difference of rotation onset between the GelSight detection and the ground truth is calculated.}\wenzhen{What is this caption part?}

\begin{figure}[t]
    \centering
    \includegraphics[width=\linewidth]{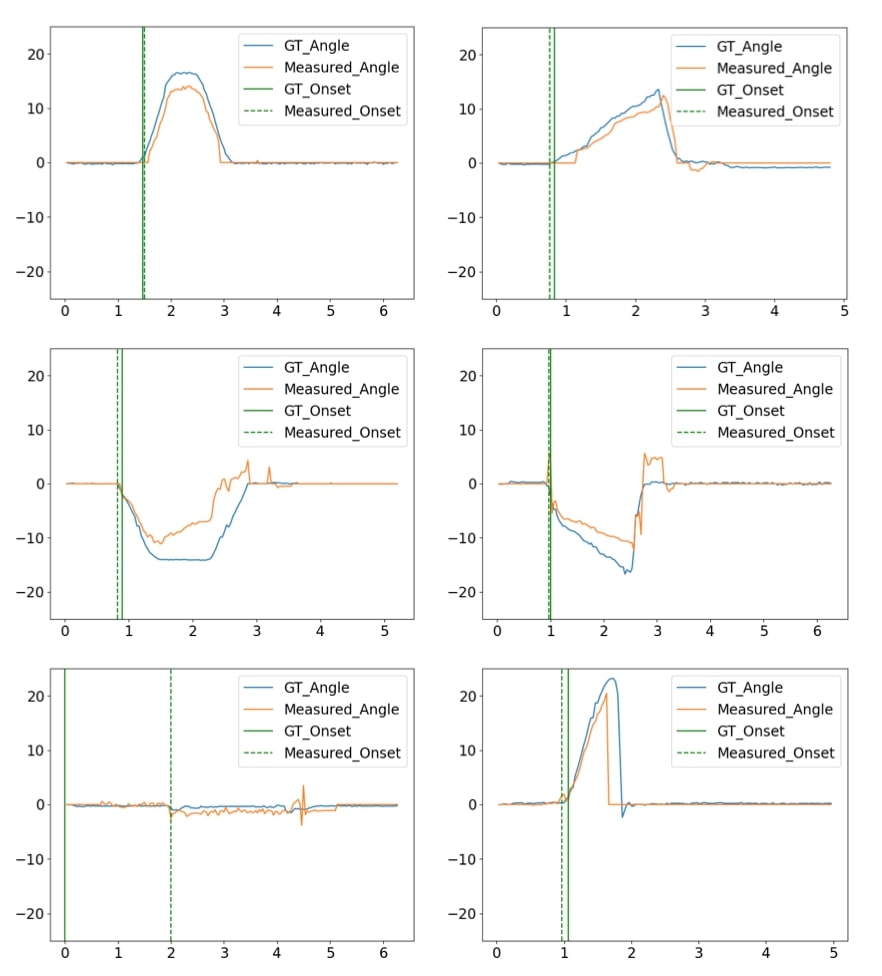}
    \caption{
    Examples of rotation angle measurement in the offline grasp experiments. x-axis represents the time (in seconds) of experiments and y-axis describes the rotation angles (in degrees). 
    % The compared results between rotation detection from GelSight sensor and ground truth rotation from external camera. 
    The cases in row 3 are one stable grasp with a small rotation angle and a rotational case with irregular-shaped contact area. 
    % Row 3 left shows a static grasp, so the rotation angles are close to zero; Row 3 right shows a failure example, where the contact area is too small and there are not enough markers in the contact area to detect rotation.
    }
    \vspace{-4mm}
    \label{fig:result}
\end{figure}

\textbf{Accuracy of rotation detection: }
We compare the measured rotation angles from the GelSight images with the ground truth of objects' rotation measured from the external camera. Some examples are shown in Fig.~\ref{fig:result}. The positive value of angles in the diagrams represent clockwise rotation, and negative values represent counter-clockwise rotation. We also evaluate the rotation onset detection delay which can be seen as vertical lines in the figures. 
% We can tell our algorithm can detect the rotation starting time and the rotation angles closely to the ground truth values from the figures.

\begin{figure}[!htbp]
    \centering
    \includegraphics[width=\linewidth]{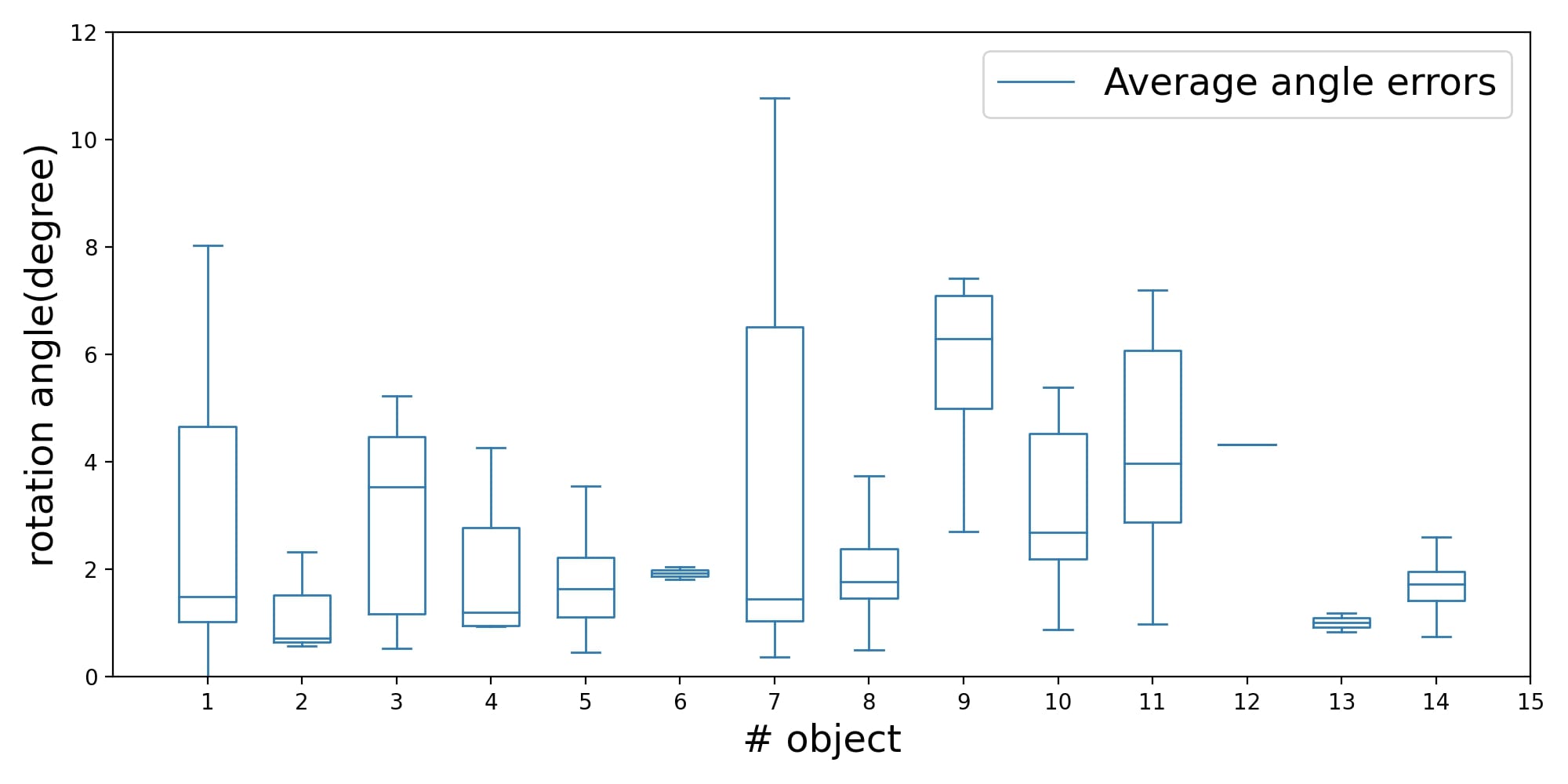}
    \caption{Rotation angle error in offline experiments. The absolute angular difference between the GelSight measurement and the ground truth value is calculated for each frame in the lifting phase, and the mean of them represents the average angle error.}
    \label{fig:angle_all}
    \vspace{-4mm}
\end{figure}

\begin{figure}[!htbp]
    \centering
    \includegraphics[width=\linewidth]{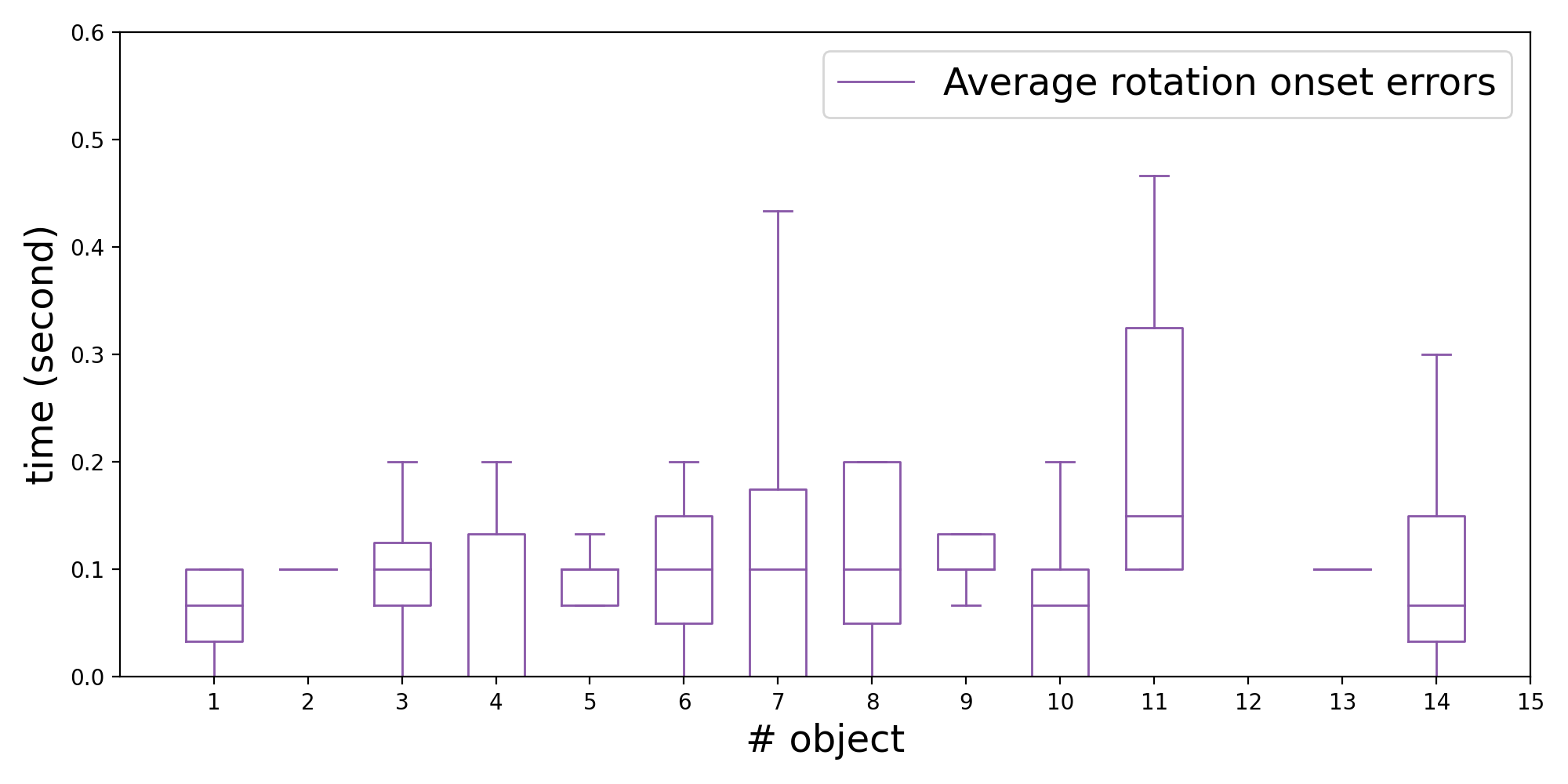}
    \caption{Rotation onset time delay in  offline experiments. The absolute time difference of rotation onset between the GelSight detection and the ground truth is calculated.}
    \label{fig:onset_all}
     \vspace{-2mm}
\end{figure}

To numerically evaluate all rotation measurement results, we estimate the mean of the absolute error between the ground truth and the measured angle for each lifting frame in each grasp trial. The lifting frame starts from the rotation onset time stamp and ends when the object detaches from the tactile sensor. The overall average error in rotation angle measurement is 3.96$^{\circ}$. However, from our observation during experiments, for cases where the rotation angles are large, the angles measured are smaller than the ground truth values. This comes from the rotational slip occurring between the contact of the gel's surface and the object. This can also be attributed to the adhesion between the gel and the object, which impedes the rotation. Unfortunately, due to the design of the sensor, these two scenarios are unavoidable. Therefore, we also evaluate the angle errors for the rotation angles under 10$^{\circ}$ to eliminate these effects. For rotation angles under 10$^{\circ}$, the average angle error is 2.69$^{\circ}$. The statistics of average angle error for all experiments can be seen in Fig.~\ref{fig:angle_all}. When contour tracking is used in angle measurement, the average angle error is 3.33$^{\circ}$ for the general case and 2.33$^{\circ}$ for the case of rotation measurement under 10$^{\circ}$. As the numbers imply, both contour tracking and marker motion tracking give similar performance in measuring the rotation angle.  Moreover, we evaluate the average time delay for rotation onset detection, as in Fig.~\ref{fig:onset_all}. The overall error is 4.22 frames, which is around 0.14 seconds.

%Comparing the two different methods proposed in this work--tracking the marker's motion patterns versus tracking the contour of the contact area--the numerical results with contour tracking are 3.33$^{\circ}$ for the average error of rotational angle measurement and 2.33$^{\circ}$ for the average error of rotation angle measurement under 10$^{\circ}$. 
%The errors are slightly lower than the overall angle measurement's error, so we would consider these two methods work equally well.
% Since the rate of Gelsight image collection is 30 fps, we set 1 second which is 30 frames as threshold to evaluate how quick our rotation onset detection is. From the diagram, most cases are under 1 second which means the reaction to rotation is quick. 

For classification between stable and rotational cases, 
% the results are shown in Table.~\ref{table:offline_result}. 
in general, 134 cases out of 142 cases are classified correctly, where the success rate is 94.37\%. Among them, 41 successful classifications out of 44 are stable cases with a success rate of 93.18\%, and 93 successful classifications out of 98 are rotational cases with a success rate of 94.90\%.
% \begin{table}[]
% \begin{tabular}{|l|l|l|l|}
% \hline
%                         & Total cases & Rotational cases & Stable cases \\ \hline
% Total \#                & 142         & 98               & 44           \\ \hline
% Corrected classified \# & 134         & 93               & 41           \\ \hline
% Accuracy                & 94.37\%     & 94.90\%          & 93.18\%      \\ \hline
% \end{tabular}
% \caption{Results for offline experiments.}
% \label{table:offline_result}
% \end{table}

Most failures come from the imperfect grasping where the contact area is located at the margin of the image. As a result, the contact is only partially visible, and even disappears after lifting, making it hard to track either the markers or the contours.
% Most failures come from small contact area or irregular object shape, which leads to irregular contact area. Curvatures can result in random marker motions on the gel, confusing the algorithm to classify unstable rotation and stable grasp. In such cases, the tracking of markers is not sufficient for accurate rotation detection. One typical failure case when the cup (object 4 in Fig.~\ref{fig:robot}) is grasped with its handle is shown in the last plot in Fig.~\ref{fig:result}. However, the results do show the algorithm's robustness in dealing with rotations of flatter objects.

% , but it is challenging to implement objects with complex shapes with curvatures. The curvatures can result in random marker motions on the gel, confusing the algorithm to classify the rotational and stable grasp.
\subsection{Closed-Loop Regrasping Experiment}
% We show how the proposed rotation detection algorithm can lead the robot to a stable grasp pose by performing closed-loop regrasping experiments using feedback from the GelSight sensor.

\begin{figure}[!htbp]
    \centering
    \includegraphics[width=\linewidth]{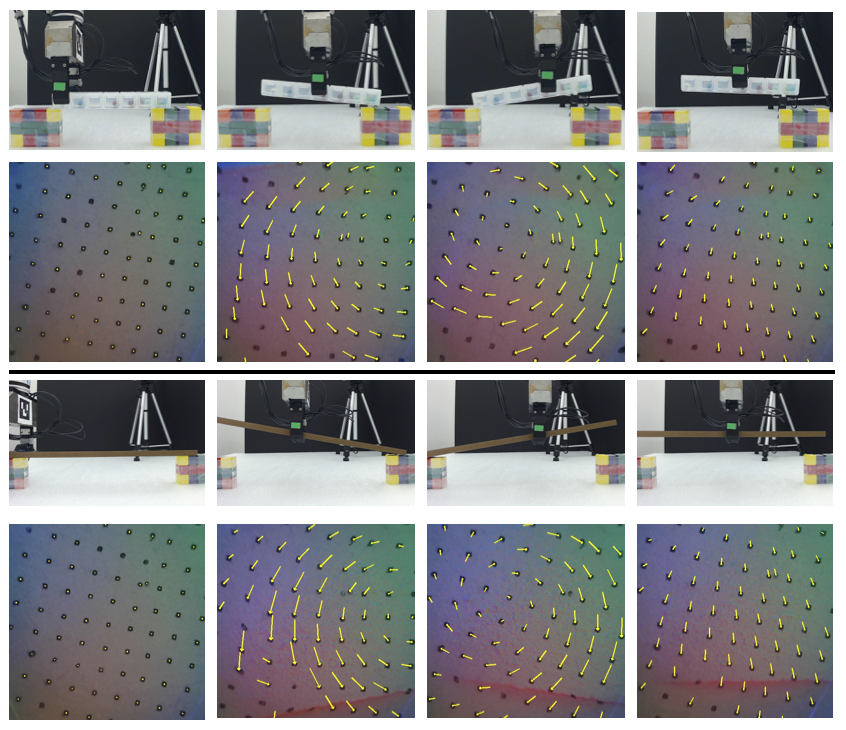}
    \caption{Closed-loop regrasping experiments. The robot detects rotation from GelSight images when lifting the object, and then adjust grasping location along the object's principal axis. The first row shows the external side view of grasping, and the second row shows the detected rotational/stable grasping from Gelsight.}
    \label{fig:location}
     \vspace{-2mm}
\end{figure}

We follow the closed-loop regrasp framework using adaptive step size as proposed in \ref{closedloopcontrol}. For each object, the robot starts near the geometric center of the object. The robot grasps the object at each grasp location and attempts to lift the object to a maximum height of 5 cm. Simultaneously, the rotation measurement algorithm analyzes each frame from the GelSight sensor. If it detects rotation greater than 5$^{\circ}$, it raises a preemptive signal to the robot through ROS. The robot then places the object back on the table. If no preemptive signal is sent till the maximum height of 5 cm, the robot then waits for 3s to check for rotation. If rotation is detected in either of the above two ways, the robot's step size is updated according to the control framework proposed in \ref{closedloopcontrol}. However, if rotation is not detected at this stage, the robot further lifts the object by 3 cm and holds it for 5s to indicate that a stable grasp pose has been achieved. The process is illustrated in Fig.~\ref{fig:location}. We manually label whether rotation occurred in a particular grasp. For a single object, we define a grasping loop as the object being grasped at multiple locations until it reaches the stable grasp pose. We conducted a total of 109 closed-loop regrasping experiments on 18 objects, at gripping speeds varying from 40-50 mm/s and maximum gripping force varying from $\sim$30-60 N. The rotation detection algorithm takes $\sim$0.045 s to process each frame while the manipulation happens.
% The robot grasps an object at a location, places it back if it gets significant rotation feedback from the sensor. It then proceeds to the next grasp location along the centroidal axis of the object and repeats the lift operation. It keeps grasping the object with incremental steps until a stable grasp is detected as illustrated in Fig.~\ref{fig:location}. We manually label whether rotation occurred in a particular grasp. For a single object, we define a grasp loop as the object being grasped at multiple locations until it reaches the stable grasp pose.

\begin{figure}[!htbp]
    \centering
    \includegraphics[width=\linewidth]{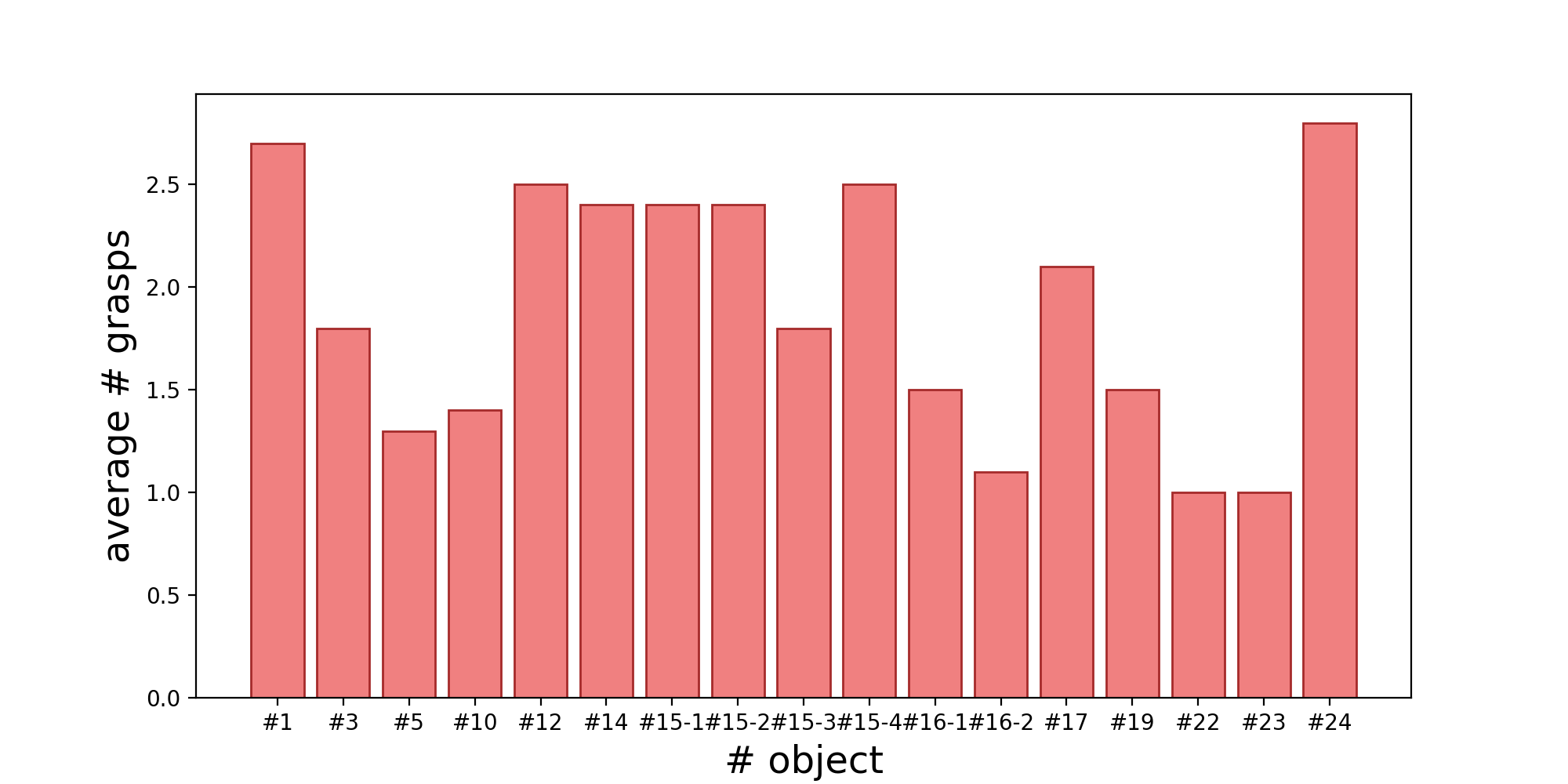}
    \caption{The average number of grasps for each object in the closed-loop experiments. The x-axis represents the object number in Fig.~\ref{fig:objects}, including different configurations of mass distributions of some objects. The y-axis represents the average number of grasps.}
    \label{fig:regrasp}
     \vspace{-2mm}
\end{figure}

\textbf{Results: }
% We conducted a total of 87 closed-loop regrasping experiments on 20 objects at gripping speeds varying from 30-50 mm/s and maximum gripping force varying from 10-50 N. The rotation detection algorithm takes $\sim$0.045 s to process each frame while the manipulation happens. \wenzhen{It's better to move those setup of speed and force number before the 'result' part}
In 105 out of the 109 experiments, the algorithm could drive the robot towards the object's centroid and avoid rotations above a set threshold of 5$^{\circ}$, with a success rate of 96.3\%. In 3 cases, the algorithm misclassified rotation as a stable grasp pose, and in 1 case a stable grasp pose as rotation. The primary reason for this failure is that the contacted area is small in the GelSight sensor's FOV during the grasp resulting in very few contact markers being detected. In cases like a pills box with uneven distribution of marbles as shown in Fig. \ref{fig:location}, our proposed method can drive the robot to a stable grasp pose in 2 regrasps on an average. Also, for objects like a spatula or vernier callipers(13, 22 in Fig. \ref{fig:objects}) 
% \wenzhen{point out the object number in Figure 7}
with uneven mass distribution, our method can detect the centre of gravity in 1.5 regrasps on average. These objects are difficult to grasp stably using vision information alone due to their uneven mass distributions; however, our method proves to be efficient in detecting center of gravity for them. On average, the robot took one regrasp to reach the center of gravity for all the experimental objects. The average number of grasps for each object in the closed-loop experiments can be seen in Fig.~\ref{fig:regrasp}. It took a maximum of 3 regrasps to reach the stable grasp pose across the set of objects.

\section{CONCLUSIONS}
% \wenzhen{If you don't have anything to discuss here, you should have 'discussion' in the title}
Rotational displacement during grasping is a common grasp failure, in which the robot grasps an object at a location far from its center of gravity. In this paper, we present an approach to detect rotational grasp failure from tactile sensing and a closed-loop regrasp 
% \wenzhen{you could use re-grasp or regrasp, but it should be consistent across the paper}
system to stabilize the grasp. The method is based on measuring the rotation angle of the soft tactile sensor's surface from tactile images, which directly correlates to the torque at the grasp surface. With the feedback of the rotation detection from tactile sensing, the robot attempts multiple regrasps and reaches a stable grasp location in the end to prevent huge rotation from happening.

% We have shown the success rate as 92.14\% in offline experiments and 90.8\% in regrasping experiments to detect rotation early and prevent grasp failures. \wenzhen{Those technical-level discussion should be in the experiment result part}However, there is space for improvement. The results show our approach's robustness to flat objects, but it is challenging to implement objects with complex shapes with curvatures. The curvatures can result in random marker motions on the gel, confusing the algorithm to classify the rotational and stable grasp.
% \Wtext{In future, we plan to extend our control framework to objects of irregular shapes and potentially use multi-fingered grippers and explore how they can give better tactile sensing capabilities. } 

In the future work, we plan to explore how learning based techniques can perform on the similar task of rotation measurement. In order for closed-loop regrasp control to work on objects of varied dimensions and shapes, we plan to include object detection and size estimation from external visual equipment and guide grasping. Also, we wish to explore control frameworks for multi-fingered robotic arms equipped with tactile sensors using our proposed method for measuring rotation. We believe higher camera fps and lower latency can help in scaling our proposed method to industrial applications. 

% \vspace{5mm}
%  \vspace{-2mm}
% In our future work, we will further analyze the generalization of rotation detection, such as with grasping complex-shaped objects or tracking loss situations.
% \wenzhen{Why do you say that? Does the current method fail with some type of objects?}And we will aim to improve the rotation detection's robustness to handle more general cases.

% Instead of tracking markers and analyzing markers' motion, we will consider the geometry of the contact surface and the surface's motion to capture more information about rotation patterns and improve rotation detection's robustness.

% \addtolength{\textheight}{-10cm}   % This command serves to balance the column lengths

%%%%%%%%%%%%%%%%%%%%%%%%%%%%%%%%%%%%%%%%%%%%%%%%%%%%%%%%%%%%%%%%%%%%%%%%%%%%%%%%

%%%%%%%%%%%%%%%%%%%%%%%%%%%%%%%%%%%%%%%%%%%%%%%%%%%%%%%%%%%%%%%%%%%%%%%%%%%%%%%%

% TODO

% \section*{ACKNOWLEDGMENT}
% TODO

%%%%%%%%%%%%%%%%%%%%%%%%%%%%%%%%%%%%%%%%%%%%%%%%%%%%%%%%%%%%%%%%%%%%%%%%%%%%%%%%
\bibliographystyle{bib/IEEEtran}
\bibliography{bib/IEEEabrv,bib/ref}

\end{document}